\documentclass[10pt,twocolumn,letterpaper]{article}
\usepackage{cvpr}
\usepackage{times}
\usepackage{epsfig}
\usepackage{graphicx}
\usepackage{amsmath}
\usepackage{cite}
\usepackage{amssymb}
\usepackage{caption}
\usepackage{subcaption}
\usepackage{multirow}
\usepackage{color}
\usepackage[pagebackref=true,breaklinks=true,letterpaper=true,colorlinks,bookmarks=false]{hyperref}

\newcommand{\ntr}{n}
\newcommand{\TT}{\mathcal{T}}

\newcommand{\LL}{\mathbf{L}}

\newcommand{\gfop}{\mathrm{MOP}}

\newcommand{\AT}{\mathbf{A}}

\newcommand{\st}{\mathrm{subject \text{ } to}}

\definecolor{lightgray}{gray}{0.9}

\newcommand{\A}{\mathbf{A}}

\newcommand{\footnoterecall}[1]{
}

 \cvprfinalcopy 

\def\cvprPaperID{1172} 

\ifcvprfinal\fi
\begin{document}

\title{Learning to Segment Moving Objects in Videos}

\author{Katerina Fragkiadaki\\
University of California, Berkeley\\
{\tt\small katef@berkeley.edu}
\and
Pablo Arbel\'aez \\
Universidad de los Andes, Colombia \\
{\tt\small pa.arbelaez@uniandes.edu.co}
\and
Panna Felsen \\
University of California, Berkeley\\
{\tt\small panna@eecs.berkeley.edu}
\and
Jitendra Malik \\
University of California, Berkeley\\
{\tt\small malik@eecs.berkeley.edu}
}
%

\maketitle
\thispagestyle{empty}


 \begin{abstract}
We segment  moving objects in 
videos by ranking 
spatio-temporal segment  proposals according to ``moving objectness'';  how likely they are to contain a moving  object.  
In each video frame, we compute segment proposals  
using multiple figure-ground segmentations on per frame motion  boundaries. 
We rank them with a  
Moving Objectness Detector 
trained on  image and motion fields to detect moving objects and 
discard over/under segmentations or background parts of the scene. 
We extend the top ranked segments  into   spatio-temporal tubes using random walkers on  motion affinities of dense point trajectories.   
Our final tube ranking consistently  
outperforms previous segmentation methods in the two largest video segmentation benchmarks currently available, 
for any number of proposals.  
Further, our per frame moving object proposals increase  the  detection rate up to 7\% over  previous state-of-the-art static  proposal methods.  
\end{abstract}

 \section{Introduction}

Proposal of regions likely to contain objects and  classification using  convolutional neural networks 
\cite{DBLP:journals/corr/GirshickDDM13} is currently 
the dominant paradigm for object detection in static images. 
Empirically,  region-CNNs have shown excellent  performance against sliding window classifiers  \cite{DBLP:journals/corr/HeZR014} that often cannot afford to enumerate all possible bounding boxes in an image, or Markov Random Field pixel classifiers that make independence assumptions regarding the organization of pixel labels, and cannot distinguish closeby instances of the same object class  \cite{Gould:2008:MSR:1416831.1416841,10.1109/TPAMI.2012.231}.   
In this paper, we propose a similar paradigm for detecting moving objects in videos by introducing motion based object proposals and  a moving objectness ranker.  We present large quantitative advances over previous multiscale segmentation and trajectory clustering methods, as well as proposal generation methods that do not consider motion boundaries or moving objectness as described in this work. 

\begin{figure}[t]
\begin{center}
\includegraphics[trim=0in 0in 0in 0in, scale=0.39]{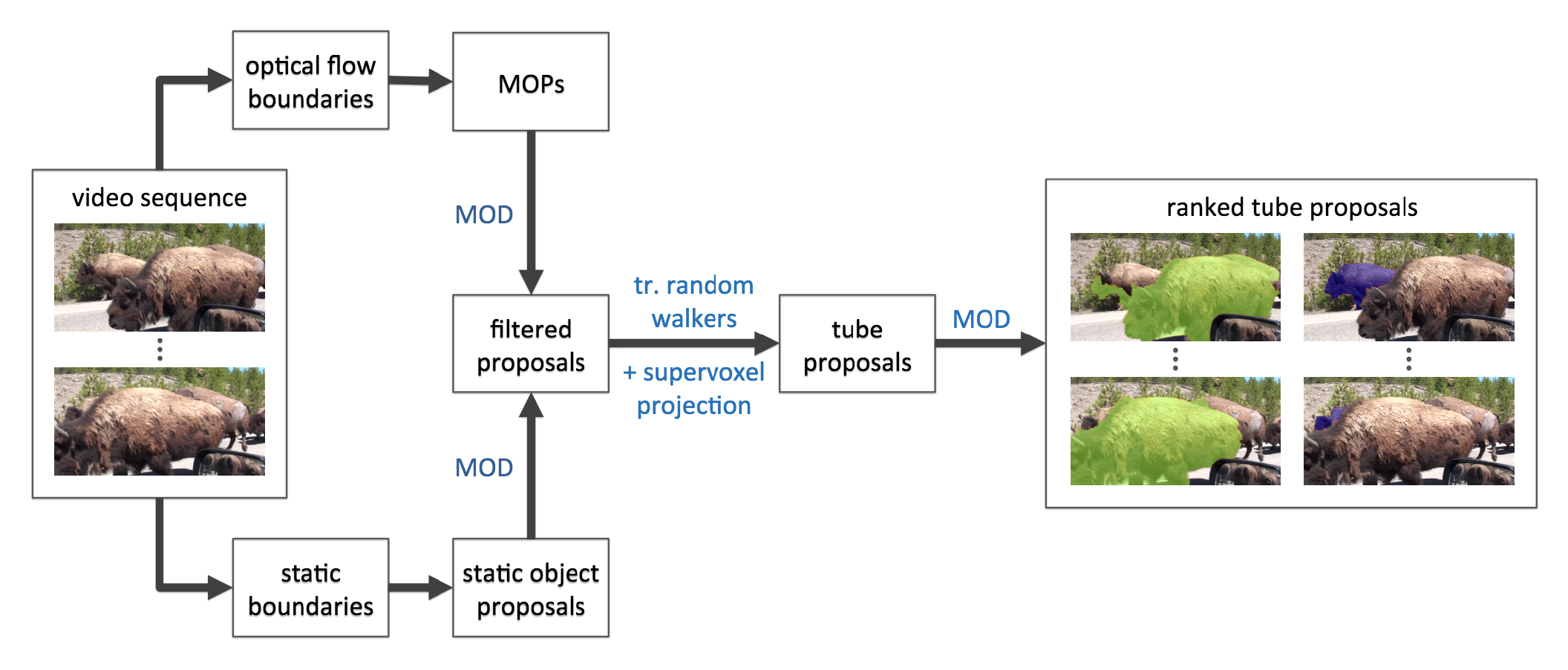}
\end{center}
\caption{\textbf{Overview.}   
We generate a set of region proposals in each frame  using multiple segmentations on optical flow and static boundaries, we call them per frame Moving Object proposals (MOPs) and static proposals.  
A Moving Objectness Detector (MOD) then rejects proposals on static background or  obvious under or over segmentations. 
The filtered proposals are extended  into spatio-temporal pixel tubes using dense point trajectories.   
Finally, tubes are ranked by our MOD using score aggregation across their  lifespans.
}
\label{fig:overview}
\end{figure}

 We propose a method that segments moving objects in monocular uncalibrated videos by  object proposal generation from multiple segmentations on motion boundaries  and ranking  with  a ``moving objectness'' detector.   
In each frame, we extract motion  boundaries by  applying a learning based boundary detector 
on the magnitude of optical flow.  
The extracted motion boundaries establish pixel affinities for  multiple figure-ground segmentations 
that generate a pool of segment proposals, which we call per frame Moving Object Proposals (MOPs). 
MOPs increase the object detection rate by 7\%  over state-of-the-art static segment proposals and     
demonstrate the value of motion for object detection in videos.  We extend per frame MOPs and static proposals into space-time tubes 
using constrained segmentation on dense point trajectories. 
The set of proposals is ranked with a  ``Moving Objectness'' Convolutional Neural Network Detector (MOD)
trained 
from  image and optical flow fields to detect moving objects and discard over/under segmentations 
and static parts of the scene. This ranking ensures  good object coverage even with a very small number of proposals. 
An overview of our approach is shown in Figure \ref{fig:overview}.

\begin{figure*}[ht]
\begin{center}
\includegraphics[trim=0in 2in 0in 1in, scale=0.15]{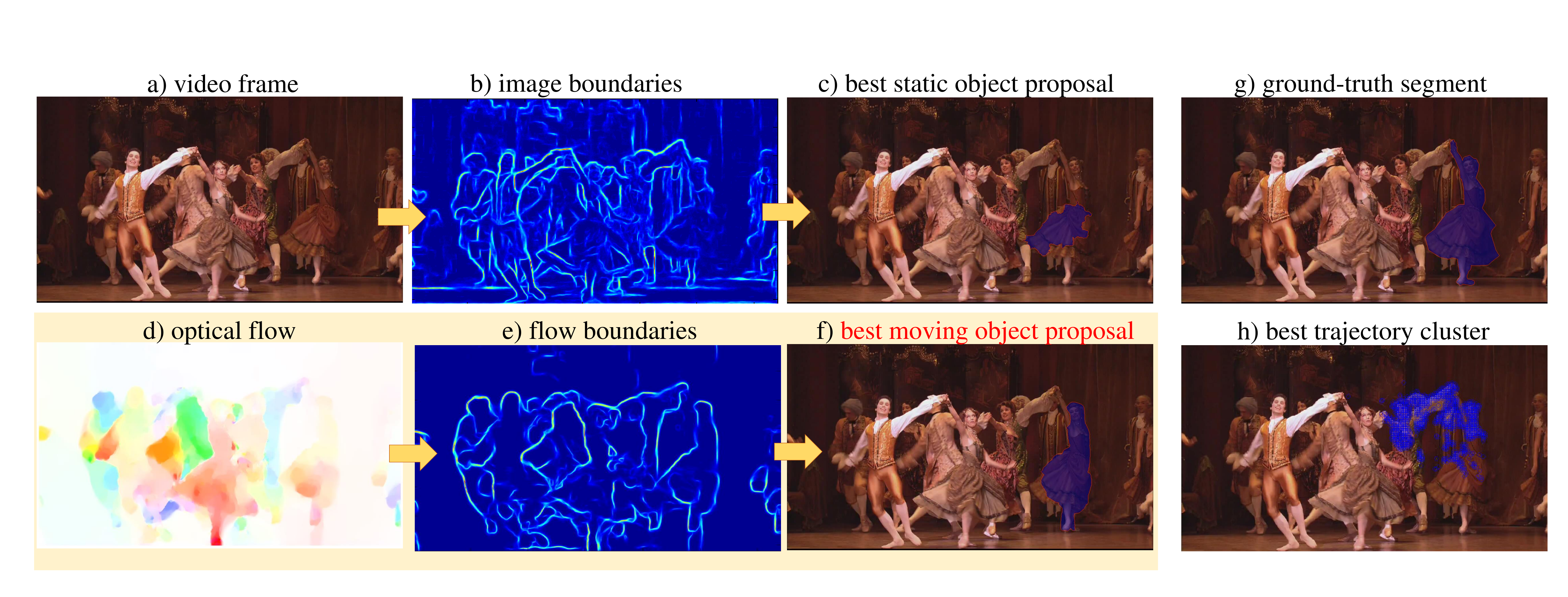}
\end{center}
\caption{\textbf{Per frame moving object proposals (MOPs). } 
Static segment proposals (c)  of \cite{gop} fail to capture the dancer as a whole due to internal clothing contours (b). Trajectory clustering \cite{springerlink:10.1007/978-3-642-15555-0_21}  fails to segment the dancer due to  frequent pixel occlusions and dis-occlusions under articulation and  trajectories being too short (h).  Flow boundaries (e) supress internal edges, and segmentation using the related pixel affinities correctly delineates the dancer (c).
}
\label{fig:gfop}
\end{figure*}

We use optical optical flow boundaries directly as input to segmentation, without combining them with static boundaries; we obtain  a diverse set of object proposals by computing grouping from RGB and motion edges separately. 
In contrast, many researchers have tried combining optical flow with static boundaries in order to
improve  boundary detection \cite{Stein_2007_5836,Bro11c}, 
with  only moderate success so far \cite{GalassoCipollaSchieleACCV12,Bro11c}. This is primarily due to optical flow misalignments 
with true object boundaries:  flow ``bleeds'' across occluding contours to the background  \cite{Thompson98exploitingdiscontinuities} because background pixels mimic the motion of the nearby foreground, as  shown in Figure \ref{fig:gfop}.   Works of \cite{Stein_2007_5836,Bro11c,oneata:hal-01021902} attempt to handle  bleeding by changing the  strength 
of static boundary fragments according to the flow content of the adjacent image regions. They are 
upper-bounded by the performance of the static boundary detector.  
For high thresholds, many boundaries are missed with no hope to be recovered. For low thresholds, overwhelming image clutter causes  regions to be too small for the flow to be aggregated effectively to fight  ``bleeding'' \cite{Bro11c}. We bypass the flow bleeding problem altogether by directly supplying slightly mis-aligned flow boundaries as input to segmentation. 

We extend per frame segments to spatio-temporal tubes using random walkers on dense point trajectory motion affinities. 
Motion is an opportunistic cue  as objects are not constantly in motion \cite{5995366}.  
 At  frames when they are static, there are no optical flow boundaries  and  MOPs miss them.   Constrained trajectory clustering propagates the segmentation from   ``lucky'', large motion frames, to  frames with little or no motion. We then map trajectory clusters to pixel tubes   according to their overlap with supervoxels.

Our Moving Objectness Detector (MOD) learns the appearance of moving objects from a set of training examples.    
It filters the otherwise exploding number of per frame segment proposals and ranks the final set of spatio-temporal tubes. The multimodal class of  moving objects is represented with  a dual pathway CNN architecture on both RGB and motion fields; its neurons capture parts of fish, people, cars, animals etc. and exploit the appearance similarities between them, e.g., many animals have four legs. The proposed MOD outperforms hand-coded center-surround saliency  and other competitive multilayer objectness baselines \cite{Hoffman14Lsda,DBLP:conf/cvpr/AlexeDF10}. 

Our method bridges the gap between motion segmentation and tracking methods.   
Previous motion segmenters \cite{XuCoCVPR2012,OB12} operate ``bottom-up'', they exploit color or motion cues without using a training set of objects. Previous  trackers \cite{5459278,5740927}  use an object detector (e.g., car or pedestrian detector) to cast attention to the relevant parts of the scene. We do use a training set for learning the concept of a moving object, yet remain agnostic to the exact  object classes present in the video.

In summary, our contributions are:
\begin{itemize}
 \item Moving object proposals from multiple segmentations on optical flow  boundaries. 
 \item A  moving objectness detector for ranking per frame segment and tube proposals.
 \item Random walks in a trajectory motion embedding for extending  per frame segments into spatio-temporal trajectory clusters.  
\end{itemize}

We test our method on the two largest video segmentation benchmarks currently available: Moseg \cite{springerlink:10.1007/978-3-642-15555-0_21} and VSB100 \cite{Galasso_2013_ICCV}. 
Our goal is to maximize Intersection over Union (IoU)  of our spatio-temporal tubes with the ground-truth  objects using as few tube proposals as possible. This is equivalent to the standard performance metric for segment proposal generation in the static domain \cite{APBMM2014,gop}. 
In each video, 55-65\% of ground-truth objects are captured in the challenging VSB100 benchmark using 64-1000 tube proposals,  outperforming competing approaches of  \cite{OB11,XuCoCVPR2012,Galasso_2013_ICCV}.  
We empirically show our method 
  can handle  articulated objects and crowded video scenes,  which are challenging cases for existing methods and baselines. 
Our code is available at \texttt{www.eecs.berkeley.edu/}$\sim$ \texttt{katef/}.

\section{Related work}\label{sec:relatedwork}

We can categorize previous methods based on the   information they assume regarding the objects  in the video   into:  
 i) top-down tracking methods,   and ii)  bottom-up video segmentation methods. 
 Tracking methods take advantage of category specific detectors to focus on the relevant parts of the scene,  
 e.g.,  pedestrian or car trackers \cite{5459278,5740927}.  
 Video segmentation methods are oblivious to  object categories. 
Works of \cite{XuCoCVPR2012,GrundmannKwatra2010} group pixels based on  color and/or optical flow similarity and  produce multiscale spatio-temporal segmentation maps. Each spatio-temporal superpixel is called a supervoxel. Work of \cite{Galasso_2013_ICCV} presents state-of-the-art results in VSB100 dataset by smoothing in time superpixels  from multiscale static boundary maps using optical flow. 
Works of \cite{springerlink:10.1007/978-3-642-15555-0_21,OB12,OB11}  cluster dense point trajectories \cite{ECCV2010GPU}  using long range trajectory motion similarities. 
They have shown excellent results on benchmarks of mostly rigid  objects. Work of \cite{OB11}  maps trajectory clusters to pixels with a multiscale Markov Random Field on per frame superpixels. 
Work of \cite{OB12} deals with trajectory sparsity by considering higher order affine models for establishing trajectory affinities.   Though  methods of \cite{springerlink:10.1007/978-3-642-15555-0_21,OB12,OB11} focus on obtaining a \textit{single} trajectory clustering \cite{springerlink:10.1007/978-3-642-15555-0_21}, we have empirically found that \textit{multiscale} trajectory clustering  effectively handles segmentation ambiguities caused by  motion variations of the objects in the scene. We will use it as an additional baseline in the experimental section. 
Many approaches have tried combining regions and point trajectories \cite{OB11,conf/eccv/RavichandranWRS12,cPalou13}.


Trajectory clusters have been shown to capture  objects for  larger temporal horizons  than supervoxels; the latter are
sensitive to  boundary strength fluctuations from frame to frame. 
However,  articulation or large motion cause frequent pixel occlusions/dis-occlusions. Trajectories are too short to be useful in that case. 

 Works of \cite{FliICCV2013,Banica_2013_ICCV_Workshops} compute multiple segment proposals per frame and link them  across frames using appearance similarity. The proposals are obtained by  multiple static figure-ground segmentations similar to \cite{conf/cvpr/LiCS10}. Work of \cite{oneata:hal-01021902} produces multiple video segments by canceling image boundaries that do not exhibit high flow strength. Both works are upper bounded by the static boundary detector. Related to us is also the work of \cite{Papazoglou_2013_ICCV} that computes object proposals directly from optical flow. They consider optical flow gradients which are  more noisy than the output of a learned boundary detector on the flow field. Further, instead of computing multiple segmentations, they compute one figure-ground hypothesis per frame by classifying pixels into figure or ground according to their spatial relationships with the flow gradients. 
 
 
 Many of the aforementioned approaches do not show results on standard benchmarks and comparison with them is difficult. In our experimental section, we compare with the popular supervoxel methods of \cite{XuCoCVPR2012,Galasso_2013_ICCV} and the state-of-the-art trajectory clustering method of \cite{OB11}, which are scalable and whose  code is publicly available. 

\begin{figure*}[ht!]
\begin{center}
\includegraphics[trim=0in 6.3in 0in 2in, scale=0.11]{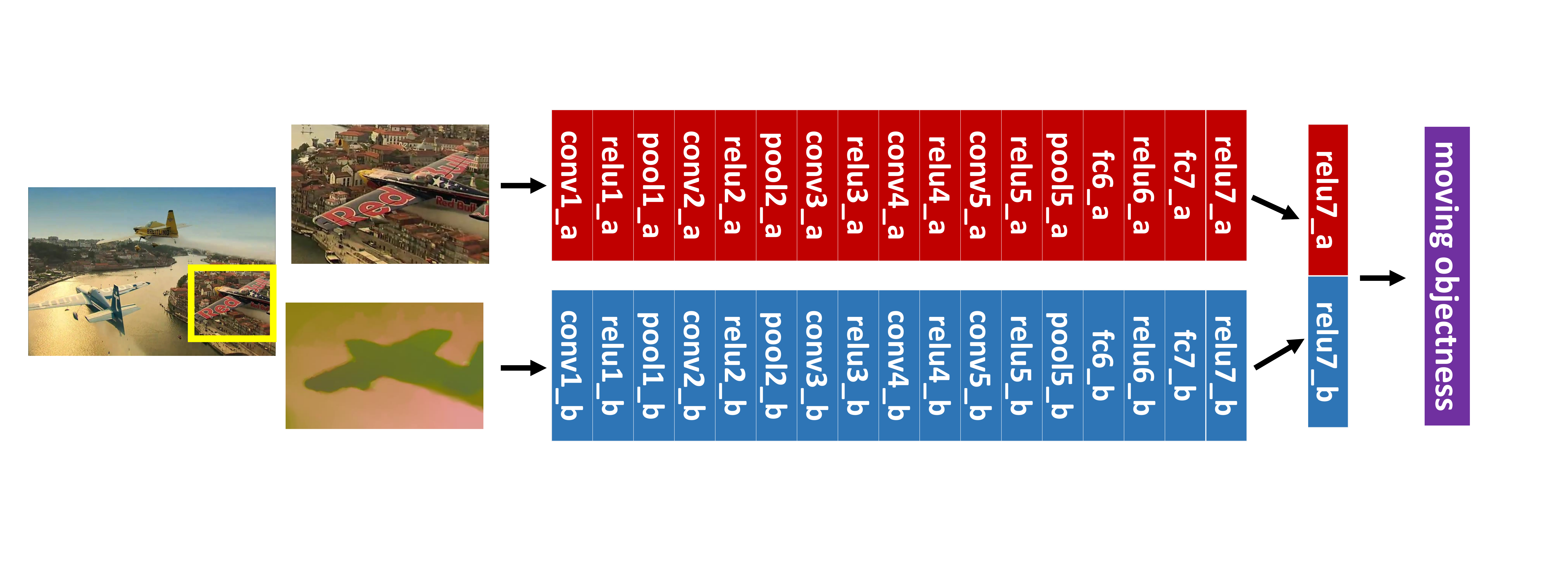}
\end{center}
\caption{\textbf{Moving Objectness Detector.} Given a bounding box of RGB and optical flow, we regress to its moving objectness score (in the interval $[0,1]$), that is, the probability it  contains a moving object versus a wrong (over or under) segmentation, or background. In the example depicted in the image the score is 1 as the input box corresponds to a correct aeroplane proposal.
}
\label{fig:mor}
\end{figure*}
 
\section{Moving Object Proposals (MOPs)}
Given a video sequence, we compute the optical flow field in each frame using the large displacement optical flow of Brox and Malik \cite{10.1109/TPAMI.2010.143}.  
Then, we compute optical flow boundaries by applying the 
state-of-the art structured forest boundary detector of Doll\'ar and Zitnick \cite{export:202540} on the magnitude of the optical flow field, which we replicate into a three channel image. Though the detector has been trained on static image boundaries of the  BSDS boundary benchmark \cite{MartinFTM01}, it effectively detects boundaries of the flow field, as shown in Figure \ref{fig:gfop}e, despite the different statistics,  e.g., the flow magnitude has many more curved corners than an RGB image of man made structures. We did not consider re-training the detector using  optical flow input  because the degree of misalignment of the flow boundaries with the true image boundaries (due to flow ``bleeding'') widely varies depending on the background texturedness, and would confuse the detector. 


We use  flow boundary maps 
to induce intervening contour based pixel affinities in the geodesic object proposal method of Kr\"ahenb\"uhl and Koltun \cite{gop}. Given a boundary map,  work of \cite{gop} computes multiple figure-ground segmentations using  randomized seed placement and superpixel classification according to shortest paths  to seeds. It has recently shown  state-of-the-art segmentation results in the PASCAL object detection benchmark. 
However, strong interior image boundaries  cause object fragmentations that persist until the saturation point of \cite{gop}, i.e., increasing the number of proposals does not improve ground-truth coverage. 
MOPs, shown in Figure \ref{fig:gfop}f, though  slightly misaligned  with the true image boundaries due to flow ``bleeding'',  improve by a margin the segmentation metrics in our video benchmarks. Optical flow effectively bridges interior boundaries due to clothing or surface marking, as motion is smooth across those, or strengthens faint cross-object contours under motion dissimilarity. 

  \section{Moving Objectness Detector}
 We  train a Moving Objectness Detector (MOD) using a CNN \cite{LeCun:1989:BAH:1351079.1351090}  with a dual-pathway  architecture operating on both  image and flow fields, shown in Figure \ref{fig:mor}. For the flow channel, we supply a 3 channel image  containing scaled $x$ and $y$ displacement fields and  optical flow magnitude. 
The architecture of each network stack is similar to Krizhevsky \textit{et al.} \cite{conf/nips/KrizhevskySH12}:  
assume $C(k,N, s)$ is a convolutional layer with kernel size $k\times k$, $N$ filters and a stride of $s$, $P(k,s)$ a max pooling layer of kernel size $k\times k$ and stride $s$, $N$ a normalization layer, $RL$ a rectified linear unit, $FC(N)$ a fully connected layer with $N$ filters and $D(r)$ a dropout layer with dropout ratio $r$. The architecture of each stack is as follows: $C(7,96,2)-RL-P(3,2)-N-C(5,384,2)-RL-P(3,2)-N-C(3,512,1)-RL-C(3,512,1)-RL-C(3,384,1)-RL-P(3,2)-FC(4096)-RL-D(0.5)-FC(4096)-RL$. 
The  relu7  features of the image and flow stacks are concatenated and a final layer  regresses to intersection over union of the input bounding box with the ground-truth segments.   
 
 We initialize the  weights in each of the two network stacks using the 200 object category  detection network of  \cite{DBLP:journals/corr/GirshickDDM13}, trained on the Imagenet detection task from RGB images.  Many moving object categories are well represented in the Imagenet training set. We also expect the detection network of \cite{DBLP:journals/corr/GirshickDDM13}, in comparison to the classification network of \cite{conf/nips/KrizhevskySH12}, to have incorporated some notion of objectness. We  finetune the network using a small collection of   boxes that capture moving objects (as well as a large set of background boxes) collected from the training sets of the VSB100 and Moseg video benchmarks. 
 We train our MOD using standard stochastic gradient descent with momentum on Caffe \cite{Jia13caffe}, a publicly available deep learning package.
 


\begin{figure*}[ht]
\begin{center}
\includegraphics[trim=0in 3in 0in 0in, scale=0.12]{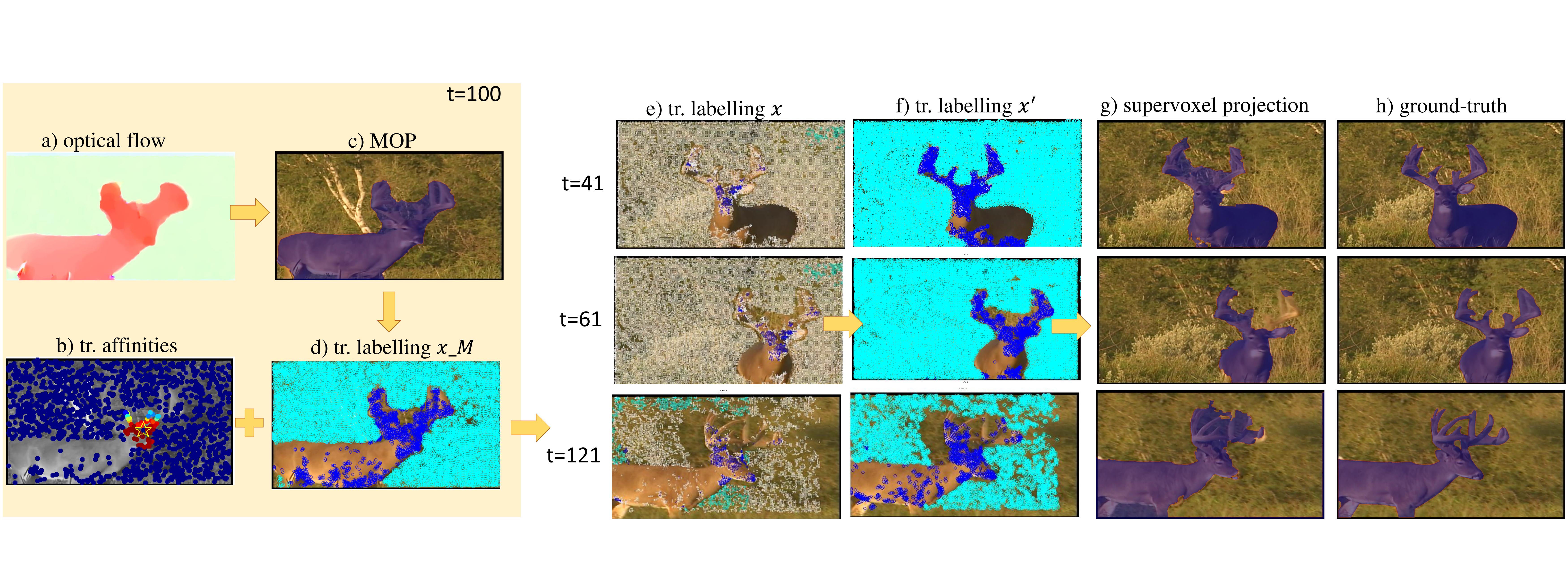}
\end{center}
\caption{\textbf{Spatio-temporal tube proposals.} 
Each per frame moving object proposal (c) labels trajectories that intersect its video frame as foreground or background (d). 
 Trajectories that do not intersect with that frame (here t=100) are unlabeled and are shown in white (e). Random walkers on trajectory motion affinities (b) effectively propagate the trajectory labels $x_M$ to the unlabeled trajectories (f). Pixel tubes are obtained from the trajectory clusters using supervoxel projection (g). 
}
\label{fig:trembedding}
\end{figure*}

\section{Tube proposal generation} \label{sec:propagation}

 We extend  per frame MOPs to spatio-temporal tubes  by propagating  pixel labels through trajectory motion affinities using Random Walkers \cite{Grady:2006:RWI:1175896.1176220}, and mapping trajectory clusters to pixels, as depicted in Figure \ref{fig:trembedding}. 



  Given a video sequence, dense point trajectories are computed by linking optical flow fields   \cite{springerlink:10.1007/978-3-642-15549-9_32}.    A trajectory terminates  when the forward-backward   consistency check fails, indicating ambiguity in correspondence. This is usually the case under pixel occlusions or dis-occlusions, or under low image texturedness.  Let $\TT$ denote the set of trajectories in the video and let $\ntr$ denote the number of trajectories, $\ntr=|\TT|$.  
  We compute pairwise trajectory  affinities $\AT \in [0,1]^{\ntr \times \ntr}$ where motion similarity between two trajectories is  
   a function of their maximum velocity difference, as proposed by \cite{springerlink:10.1007/978-3-642-15555-0_21}, and thus is robust to per frame  ambiguous motion.   We  compute affinities between each pair of trajectories that overlap in time and   are within a spatial distance of 60 pixels.  Trajectory affinities are visualized in Figure \ref{fig:trembedding}b. 

  Let $t_i$ denote the frame that  $\gfop_i$ is detected. 
  Point trajectories that intersect frame  $t_i$ are labeled as foreground or background. They are shown in Figure \ref{fig:trembedding}d in blue and light blue, respectively. Trajectories that terminate before or start after  $t_i$  are unlabeled. They are shown in white in Figure \ref{fig:trembedding}e. Let $x \in \{0,1\}^\ntr$ denote trajectory labels, $1$ stands for foreground and $0$  for background.  Let $F$ denote the foreground  and $B$  the background trajectory sets, respectively, and let $M=F \cap B$ denote the set of labeled (marked) trajectories and $U=\TT \backslash M$  the set of unlabeled trajectories. Let $\LL$ denote the trajectory un-normalized Laplacian matrix: $\LL=\mathrm{Diag}(\A\mathbf{1}_\ntr)-\A$, where $\mathrm{Diag}(y)$ stands for a diagonal matrix with vector $y$ in the diagonal. We 
  minimize the random walker cost function proposed in 
  \cite{Grady:2006:RWI:1175896.1176220}:

\begin{equation}
\label{eq:randomwalker}
 \begin{array}{cc}
  \displaystyle \min_{x}. & \quad \frac{1}{2}x^T\LL x \\
 \st & \quad x_B=0, \quad  x_F=1.
 \end{array}
\end{equation}
It is easy to show that minimizing $x^T\LL x$ is  equivalent to minimizing $\displaystyle\sum_{i,j}^n\A_{ij}(x_i-x_j)^2$ . 
We relax $x$ to take real values, $x \in [0,1]^\ntr$. 
Then Eq. \ref{eq:randomwalker} has a closed form solution given by: 
$\LL_Ux_U=-\LL^T_{MU}x_M$, where $x_U$ are the labels of the unlabeled trajectories we are seeking, and $x_M$ are the  labels of the marked trajectories. 
 We approximate computationally this closed form solution by performing a sequence of label diffusions using the normalized affinity matrix:
 \begin{equation}
 x'=\mathrm{Diag}(\A\mathbf{1}_\ntr)^{-1} \A x. 
 \end{equation}
 We have found 50 diffusions to be adequate for our radius of affinities of around 60 pixels in each frame.  
  We show in Figure \ref{fig:trembedding}f the diffused trajectory labels. 

  
 We  map trajectory clusters to pixels using a weighted average over supervoxels, superpixels that extend across multiple frames.  We compute supervoxels by greedily smoothing superpixel labels in time, similar to \cite{Galasso_2013_ICCV}. The weight of each supervoxel is its Intersection over Union (IoU)  score with the trajectory cluster. We  threshold the weighted average to obtain a binary spatio-temporal segmentation for each trajectory cluster,  shown at Figure \ref{fig:trembedding}g: the deer has been fully segmented from its background.  Notice that sharp boundaries have been recovered despite the misaligned boundaries of the generating MOP in   Figure \ref{fig:trembedding}c. Also, image parts sparsely populated by trajectories due to low image texturedness, such as the deer body, have been correctly labeled. 
 
      \begin{figure*}[ht!]
\begin{center}
\includegraphics[trim=0.5in 0.5in 0in 0in, scale=0.17]{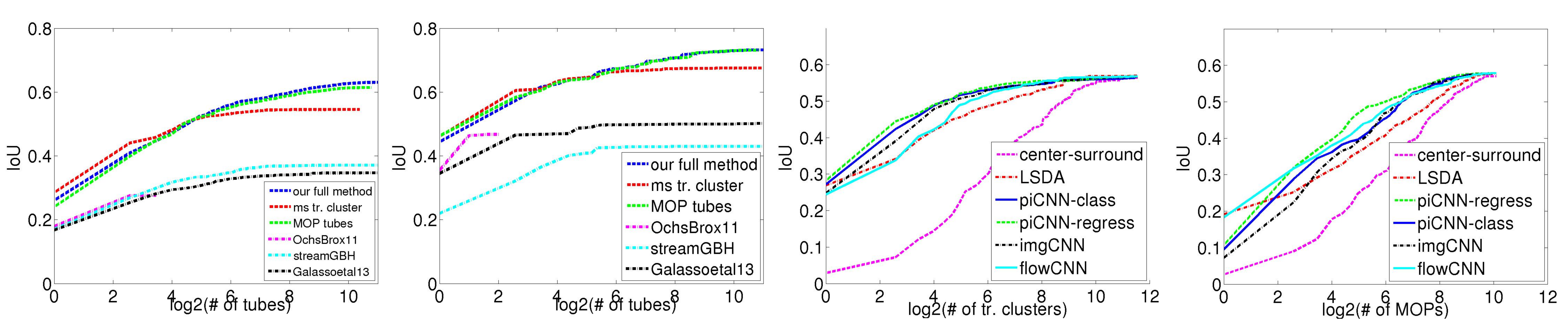}
\end{center}
\caption{ Cols 1,2: \textbf{Motion segmentation results in VSB100} (col. 1) \textbf{and Moseg} (col. 2). Our method outperforms  previous supervoxel and trajectory clustering approaches. Cols 3,4: \textbf{Ranking tube} (col. 3) \textbf{and per frame segment}  \textbf{proposals} (col. 4). Our dual-pathway CNN regressor outperforms  other CNN alternatives and hand-coded center-surround saliency. 
}
\label{fig:allcurves}
\end{figure*}

 \section{Experiments}\label{sec:experiments}
 We test our method on the two largest publicly available video segmentation benchmarks: VSB100 \cite{Galasso_2013_ICCV} and Moseg \cite{springerlink:10.1007/978-3-642-15555-0_21}. VSB100 contains 100 video sequences, 40 training and 60 testing, they are high definition  videos collected from Youtube.  
 Object motion can be very subtle or extremely articulated. Many crowded scenes are included, such as a parade, a cycling race, beach volley, ballet, salsa dancing etc. We focus on ``rigid and non-rigid motion subtasks'' of the VSB100  benchmark that concern moving object segmentation (as opposed to segmenting  static background).   The Moseg dataset contains 59 videos 
 that depict scenes from the Hollywood movie ``Miss Marple'', as well as  cars and animals, e.g., cats, rabbits, bears, camels, horses, etc. The moving objects have  distinct motion to surroundings and the scenes are relatively uncluttered, with few (one or two on average) objects per video. 
 
 First, we  benchmark our complete motion segmentation method and compare against state-of-the-art single level point trajectory clustering of \cite{OB11}, as well as the supervoxel methods of \cite{XuCoCVPR2012,Galasso_2013_ICCV}.  Our method reaches higher ground-truth coverage than previous works for any number of proposals. Second, we benchmark per frame MOPs  on static image segmentation.  
 We show that when MOPs are combined with  static segment proposals of  \cite{gop} they achieve  average best overlap, coverage and detection rates that surpass the saturation point of  static segment proposals.   
 Last, we benchmark our moving objectness detector on ranking per frame segments as well as spatio-temporal tube proposals,  and compare with alternative CNN architectures, center-surround saliency and static image objectness. 



 \paragraph{Motion segmentation}
 We compare our method with popular supervoxel methods of \cite{XuCoCVPR2012,Galasso_2013_ICCV} and the trajectory clustering and pixelization  method of \cite{OB11}. 
 For \cite{Galasso_2013_ICCV} we use our own implementation since our supervoxel computation closely follows their method. For \cite{XuCoCVPR2012} and \cite{OB11} we use code available online by the authors. 
 For both our method and the baselines we use our moving objectness detector to rank their spatio-temporal segments.  
  Score diversification has been used as in \cite{Carreira10constrainedparametric} for soft non-maxima suppression.   
   Hierarchical spatio-temporal segmentation of \cite{GrundmannKwatra2010}, distributed with the code of \cite{XuCoCVPR2012}, was not scalable enough to use in our benchmarks.

  Our MOD ranker allows us to exploit  diverse sets of tube proposals. We consider multiscale trajectory clustering as one such source, that complements our MOP tubes. Specifically, we discretize the spectral embedding of trajectory motion affinities \cite{Jianbo03multiclassspectral} for varying the number of eigenvectors. We used $50$ as the maximum number of eigenvectors used in all our experiments.   


  We show motion segmentation results on VSB100 and Moseg benchmarks in Figure \ref{fig:allcurves} columns 1 and 2, respectively. The horizontal axis denotes number of proposals used per video sequence and the vertical axis denotes the  Intersection over Union with the ground-truth spatio-temporal segments, averaged across video sequences. We score separately \textit{MOP tubes}, multiscale trajectory clusters (\textit{ms tr. cluster}) as well as their union, which is \textit{our full method}.  
  Our method outperforms previous approaches for any number of tube proposals. Multiscale trajectory clustering does not offer significant boost over MOP tubes, yet on its own is a very competitive baseline. Notice also the big difference in performance of all methods across the two datasets, indicative of the more challenging nature of VSB100 over Moseg (many non-rigid objects, subtle or articulating motion etc.).

 



 \paragraph{Static segmentation}
 We  test the performance of MOPs on   object segmentation in each frame. 
 We consider the following four widely used  static image segmentation metrics: a) \textit{Average best overlap}: the average (across all \textit{2D} ground-truth segments in our dataset) of the best IoU score of a ground-truth object with all segment proposals. b) \textit{Coverage}:  the weighted average  of IoU scores, weighted by the area of the ground-truth segments (larger segments matter more). c) \textit{Detection rate at 50\%}:the percentage of ground-truth segments that have IoU above 50\% with a segment proposal. d) \textit{Detection rate at 70\%}. It has been shown in \cite{gop} that a threshold of 70\% asks for more perceptual similarity between objects and is a better metric for object detection.   We further present \textit{anytime best} (ab) versions of a, c and d metrics, where for each ground-truth \textit{tube} (rather than per frame segment) we consider the best overlap with a segment proposal throughout its lifespan; this metric upper-bounds the performance of our MOP tubes.
 
 We show results of the proposed \textit{MOPs}, static geodesic object proposals  of \cite{gop} (\textit{GOPs}) and combined segment proposals (\textit{GOP+MOP}) in Table \ref{table:gfop}. Next to each method, we show in parentheses the number of segment proposals used.  Combining MOPs and GOPs achieves an increase of 6\% and 5\% of the detection rates at 50\% and 70\% overlap, respectively,  in the challenging VSB100 benchmark, and 5\% increase  of the detection rate at  70\% overlap in Moseg, for the same number of proposals.  This shows GOPs and MOPs are complementary, they fail and succeed at different places. The performance boost is larger in the VSB100 dataset. These numbers cannot be achieved by increasing the number of proposals in \cite{gop} which we observed reaches its saturation point at 2500 number of proposals per frame. 


\begin{table*} [ht]
\setlength{\tabcolsep}{2.5pt}
\begin{tabular}{|c|c|c|c|c|c|c|c|c|}
\cline{1-9}
\multicolumn{2}{|c|}{}        & avg best ol & coverage & det 50\% & det 70\% & avg best ol ab &  det 50\% ab & det 70\% ab\\ \hline
 \multirow{3}{*}{VSB 100} &GOP  (2715) & 53.74      & 66.84           & 60.34       & 26.12       & 65.08      &  82.6 & 48.08 \\ 
                       & MOP (873) & 46.47     & 61.3              & 47.25       & 13.85       & 57.92      & 73.75 & 29.79\\ 
                       & GOP+MOP (2659=1786+873)& \textbf{56.17} & \textbf{69.85}       & \textbf{66.48} & \textbf{31.50} & \textbf{67.15}      &\textbf{86.14} & \textbf{51.92}\\ \hline


\multirow{3}{*}{MOSEG} &GOP (2500)   &68.47      & 76.56              & 87.59       & 64.54       & 74.72         &91.94 & 79.03 \\ 
                       & MOP(839) &57.74      & 68.49              & 70.57       & 37.94       & 66.42         & 83.87 & 59.68\\ 
                       & GOP+MOP (2512=1673+839)& \textbf{69.65}      & \textbf{78.29}             & 87.59      & \textbf{70.21}       &\textbf{75.38}            &91.94 & \textbf{83.87}\\ \hline

\end{tabular}
\caption{\textbf{Static segmentation results.}  We compare geodesic object proposals (\textit{GOPs}) of \cite{gop}, per frame \textit{MOPs} proposed in this work, and a  method that considers both (\textit{GOP+MOP}).  We show in parentheses the number of proposals used in each method. 
The performance boost from combining GOPs and MOPs,  though significant in both datasets,  is  larger for VSB100 that contains heavily cluttered scenes. There, the static boundary detector often fails, and motion boundaries have a good chance of improving over it.  
}
\label{table:gfop}
\end{table*}
 
 \paragraph{Proposal ranking}
 We test our moving objectness detector on ranking  per frame MOPs and  spatio-temporal tubes produced by multiscale trajectory clustering in the VSB100 dataset.  We show the corresponding ranking curves produced by  averaging across images in the first case and across video sequences in the second in Figure \ref{fig:allcurves} at columns 3 and 4, respectively. The curves indicate how many segments/tubes are needed to reach a specific level of Intersection over Union score with the ground-truth segments/tubes.  We define the score of each tube as the \textit{sum of the scores of the bounding boxes throughout its lifespan}. We use  sum instead of average because we want to bias towards longer tube proposals. 
 
 We compare our dual-pathway  CNN regressor from image and flow fields (\textit{piCNN-regress}) against a dual pathway classification CNN (\textit{piCNN-class}), an image only  CNN (\textit{imgCNN}), a flow only  CNN (\textit{flowCNN}), our implementation of a standard center-surround saliency measure from optical flow magnitude (\textit{center-surround}) \cite{19146246},  and an objectness detector using the 7000 category detector from the Large Scale Domain Adaptation (LSDA) work of \cite{Hoffman14Lsda}.  Our dual pathway classification CNN is trained to classify boxes as positive or negatives using a threshold of 50\% of IoU, instead of regressing to their IoU score. For our LSDA baseline, we consider for each per frame segment bounding box $b$ a weighted average of the confidences of the detection boxes of \cite{Hoffman14Lsda}, where weights correspond to their intersection over union with box $b$. We have found this objectness baseline to provide a  competitive static objectness detector. 
 
  Our  dual-pathway CNN regressor  performs best among  the alternatives considered, though has close performance with the dual-pathway  classification CNN.  Our CNN networks operate on the bounding box of a segment rather than its segmentation mask. While masking the background is possible, context is important for judging over and under-segmentations.

 \paragraph{Discussion - Failure cases}
 In the VSB100 dataset, many failure cases concern temporal fragmentations. They are caused by large motion or full object occlusions. Our method as well  as our baselines would benefit from an additional linking step, where similarly looking tubes are linked across to form longer ones. To keep the method clean we did not  consider such a step. In Moseg dataset, most failure cases are due to inaccurate mapping of trajectory clusters to pixel tubes: we often slightly leak to the background, especially for animals with thin limbs, such as camels. 
 
 \paragraph{Computational time}
 The following numbers are for a single cpu. 
 Large displacement optical flow takes on average 16 secs per image. 
 Given an optical flow field, computing MOPs  takes 4 seconds on an 700X1000 image. The projection of each MOP  to the trajectory embedding takes 2 seconds for 70000 trajectories,  all MOPs can be projected simultaneously using matrix diffusion operations.  Supervoxel computation is causal and takes 7 seconds in each frame. Computing motion affinities for 70000 trajectories takes 15 seconds in each video. Our supervoxel computation, optical flow computation, MOP computation and projection are completely parallelizable. 
 
    
\section{Conclusion}\label{sec:conclusion}
We have presented a  method that segments moving objects in videos by  mutliple segment proposal generation and ranking according to moving objectness. 
Our moving object proposals  complement  static ones, 
and boost by a  margin their performance of capturing moving objects, especially in cluttered, challenging scenes. 
The proposed  moving object detector   discards over and under  fragmentations  or background parts of the scene, and provides a  ranking that allows to capture  ground-truth objects with few tube proposals per video. The proposed method   bridges the gap between video segmentation and tracking research, by exploiting training sets for learning the appearance of moving objects, yet not committing to a single object class of interest and  
by representing objects with pixel tubes instead of bounding box tracklets. 

\vspace{-0.2in}
\paragraph{Acknowledgesments}
We gratefully acknowledge NVIDIA corporation for the donation of Tesla GPUs for this research. This research was funded by ONR MURI N000014-10-1-0933.

{\small
\bibliographystyle{ieee}
\bibliography{refs}
}

\end{document}